\documentclass[runningheads]{llncs}
\usepackage[T1]{fontenc}
\usepackage{graphicx}
\usepackage{amsmath}
\usepackage{amssymb} 
\usepackage{algorithm}
\usepackage{algorithmic}
\usepackage{multirow}
\usepackage{booktabs}
\usepackage{makecell}

\begin{document}
\title{P2C: Path to Counterfactuals}
%
%
\author{Sopam Dasgupta\inst{1}\orcidID{0009-0008-3594-5430} \and
Sadaf Halim\inst{1}\orcidID{0000-0002-4152-8122} \and
Joaqu\'in Arias\inst{2}\orcidID{0000-0003-4148-311X}\and
Elmer Salazar\inst{1}\orcidID{0000-0002-3042-474X}\and
Gopal Gupta\inst{1}\orcidID{0000-0001-9727-0362}}
\authorrunning{S. Dasgupta et al.}
%
\institute{The University of Texas at Dallas, Richardson TX 75080, USA \\
\email{\{sopam.dasgupta, SadafMd.Halim, ees101020, gupta\}@utdallas.edu}
\and
CETINIA, Universidad Rey Juan Carlos, Madrid, Spain\\
\email{joaquin.arias@urjc.es}}
\maketitle              
\begin{abstract}

Machine-learning models are increasingly driving decisions in high-stakes settings, such as finance, law, and hiring, thus, highlighting the need for transparency. However, the key challenge is to balance transparency—--clarifying `why' a decision was made—--with recourse: providing actionable steps on `how' to achieve a favourable outcome from an unfavourable outcome. Counterfactual explanations reveal `why' an undesired outcome occurred and `how' to reverse it through targeted feature changes (interventions). 

~~~~Current counterfactual approaches have limitations: 1) they often ignore causal dependencies between features, and 2) they typically assume all interventions can happen simultaneously, an unrealistic assumption in practical scenarios where actions are typically taken in a sequence. As a result, these counterfactuals are often not achievable in the real world. 

~~~~We present P2C (Path-to-Counterfactuals), a model-agnostic framework that produces a plan (ordered sequence of actions) converting an unfavourable outcome to a \textit{causally consistent} favourable outcome. P2C addresses both limitations by 1) Explicitly modelling causal relationships between features and 2) Ensuring that each intermediate state in the plan is feasible and causally valid. P2C uses the goal-directed Answer Set Programming system s(CASP) to generate the plan accounting for feature changes that happen automatically due to causal dependencies. Furthermore, P2C refines cost (effort) computation by only counting changes actively made by the user, resulting in realistic cost estimates. Finally, P2C highlights how its causal planner outperforms standard planners, which lack causal knowledge and thus can generate illegal actions.

\vspace{-0.15in}
\end{abstract}
\section{Introduction}

Machine Learning systems are increasingly entrusted with high‑stakes decision-making tasks, ranging from job candidate screening to loan approvals. Yet their opaque, “black‑box” nature makes it difficult for affected individuals to understand why a particular decision was reached, especially when the outcome is undesired or unfavourable. Additionally, beyond mere explanations, people often want to know \textbf{how} they could flip a negative outcome (e.g., loan denied) into a positive one (loan approved). A solution to the above problems has been provided by counterfactuals explanations. Counterfactuals explain not only \textbf{`why'} a decision outcome was made but also \textbf{`how'} to achieve a desired outcome by specifying the necessary feature changes, i.e., interventions. However, current counterfactual approaches have limitations: (1) they often ignore causal dependencies between features, and/or (2) they typically assume that all feature changes can happen simultaneously---an unrealistic assumption in practical scenarios where actions are typically taken in a sequence. 

We introduce P2C (Path to Counterfactuals), a model‑agnostic framework that automatically generates a sequential path to a minimum cost causally compliant counterfactual explanation for any classifier—--statistical or rule‑based. P2C follows the given steps:
1) Rule Learning: Extracts interpretable decision rules from the underlying black box model and causal rules from data if the causal model is not provided. 
\textbf{Optional:} If the causal model is not provided, \textbf{Expert Verification} is used to ensure learned causal dependencies from Step 1 reflect true causation, not spurious correlations. 
2) Using the decision and causal rules, it obtains the minimum effort counterfactual $g$ that is causally consistent.
3) Uses Goal‑Directed ASP system, s(CASP), to treat counterfactual search as a planning problem, by finding a step‑by‑step path of allowable feature changes from the initial (negative) state $i$ to the counterfactual/ goal (positive) state $g$.

In P2C, the initial state $i$ satisfies the underlying classifier’s reject-condition (e.g., {\tt`?- reject\_loan(john)'} succeeds), while any goal state $g\in G$ must flip that decision (e.g., {\tt`?- not reject\_loan(john)'} succeeds). Finding a path from $i$ to $g$ becomes an ASP‑driven planning task in which every \textbf{direct action} (direct change of a feature value) and \textbf{causal action} (downstream causal changes triggered due to preceding direct action) respects causal dependencies between features. For instance, one cannot directly alter the credit score. In order to increase the credit score one must first clear debt: P2C encodes that “$no\_debt \rightarrow credit\_score\geq 620$”. This leads to a direct action of clearing debt having the causal effect of increasing the credit\_score to `$\geq$ 620'.

Unlike current counterfactual approaches that either ignore causal constraints or return an unordered set of actions, P2C provides an ordered step-by-step series of actions/interventions (direct or causal) to the counterfactual state by leveraging s(CASP)’s \cite{scasp-iclp2018} native backtracking and compact FOLD‑SE \cite{foldse} rule sets. 
Additionally, P2C produces paths guaranteed to contain only legal actions (does not directly alter \textit{credit\_score}), unlike that of a standard planner, which will take illegal actions due to lack of causal knowledge. In the next section, we review related work before detailing the P2C methodology.

\vspace{-0.2in}
\section{Background and Related Work}

\vspace{-0.1in}

\subsection{Counterfactual Reasoning}

Explanations help in understanding decisions. Counterfactuals \cite{wachter} were used to explain individual decisions by offering insights on achieving the desired outcomes. For instance, a counterfactual explanation for a loan denial might state: If John had \textit{good} credit, his loan application would be approved. This involves imagining alternate (reasonably plausible) scenarios where the desired outcome is achievable. For a binary classifier given by $f:X \rightarrow \{0,1\}$, we define a set of counterfactual explanations $\hat{x}$ for a factual input $x \in X$ as $\textit{CF}_{f}(x)=\{\hat{x} \in X | f(x) \neq f(\hat{x})\}$. This set includes all inputs $\hat{x}$ leading to different predictions than the original input $x$ under $f$. Various counterfactual methods \cite{ref_asp_cf}, \cite{ref_2_ustun}, \cite{alt_karimi}, \cite{ref_mace_1}, \cite{ref_mace_2} were proposed, however, they assumed feature independence. This resulted in unrealistic counterfactuals as in the real world, causal dependencies exist between features.

\vspace{-0.2in}
\subsection{Causality and Counterfactuals}
\vspace{-0.05in}
Causality relates to the cause-effect relationship among variables, where one event (the cause) directly influences another event (the effect). In the causal framework \cite{SCM}, causality is defined through \textit{interventions}. An \textit{intervention} involves external manipulation of $P$, which explicitly changes $P$ and measures the effect on $Q$. 
This mechanism is formalized using \textit{Structural Causal Models (SCMs)}, which represent the direct impact of $P$ on $Q$. \textit{SCMs} allow us to establish causality by demonstrating that an intervention on $P$ leads to a change in $Q$.
In \textit{SCM}-based counterfactual approaches such as \textit{MINT} \cite{ref_4_karimi_2}, capturing the downstream effects of interventions is essential to ensure causally consistent counterfactuals. 
However this is challenging: real-world domains often contain feedback (e.g., low credit score $\rightarrow$ high debt $\rightarrow$ even lower credit score) and context-dependent effects that are difficult to capture with the acyclic graphs underpinning most DAG-based SCMs. 

By explicitly modelling counterfactual dependencies, \textit{SCMs} help in generating counterfactuals. While \textit{MINT} produces causally consistent counterfactuals, it typically proposes a simultaneous set of minimal interventions/actions. When applied together, this set of interventions produces the counterfactual solution. However, executing such a set of simultaneous interventions in the real world might not be possible. The order matters as some interventions may depend on others or require different time-frames to implement. 
Sequence planning is necessary to account for such causal dependencies and practical constraints. \textit{C3G} \cite{C3G}, a counterfactual approach, also suffers from this problem of only providing a set of interventions without an order. Like \textit{MINT}, \textit{C3G} considers causal dependencies while producing counterfactuals. However, unlike \textit{MINT}, \textit{C3G} relies on Answer Set Programming to generate the counterfactuals.

\vspace{-0.15in}
\subsection{Answer Set Programming (ASP)}
\vspace{-0.1in}
\textbf{Answer Set Programming (ASP)} is a paradigm for knowledge representation and reasoning \cite{cacm-asp,gelfond-kahl}. Widely used in automating commonsense reasoning, ASP inherently supports non-monotonic reasoning allowing conclusions to be retracted when new information becomes available. This is helpful in dynamic environments where inter-feature relationships may evolve, allowing ASP to reason effectively in the presence of incomplete or changing knowledge.
In ASP, we can model the effect of interventions by defining rules that encode the relationship between variables. ASP can simulate interventions through causal rules. For example, rules in ASP specify that when $P$ is $TRUE$, $Q$ follows, and similarly, when $P$ is $FALSE$, $Q$ is also $FALSE$: $(P \Rightarrow Q)$ $\wedge$ $(\neg P \Rightarrow \neg Q)$. By changing $P$ (representing an intervention), ASP can simulate the effect of this change in $Q$, thereby capturing causal dependencies similar to \textit{SCMs}. 
Unlike DAG-based SCMs, ASP rules can natively encode cycles. A recurrent relationship---{\tt low\_credit\_score :- high\_debt.} and {\tt high\_debt :- low\_credit\_score.}---captures the mutual reinforcement between credit score and debt that a DAG based \textit{SCM} would forbid.
Conditional dependencies are equally succinct; e.g., {\tt loan\_approved :- high\_income, not low\_credit\_score.} states that high income secures approval only if credit score is not low. This expressiveness allows ASP to model intricate, sometimes cyclic feature interactions encountered in real datasets, complementing SCMs where strict acyclicity is too restrictive.

To execute ASP programs efficiently, \textbf{s(CASP)} \cite{scasp-iclp2018} is used. It is a goal-directed ASP system that executes answer set programs in a top-down manner without grounding. Its query-driven nature aids in commonsense and counterfactual reasoning, utilizing proof trees for justification. 
To incorporate negation, s(CASP) adopts \textit{program completion} as shown \cite{baral}, turning ``if'' rules into ``if and only if'' rules: $(P \Rightarrow Q)$ $\wedge$ $(\neg P \Rightarrow \neg Q)$. 
Through these mechanisms, ASP in \textit{P2C} provides a novel framework for generating realistic counterfactual explanations in a \textit{step-by-step} manner.
\vspace{-0.1in}
\subsection{FOLD-SE}
\vspace{-0.1in}
A good example of a \textbf{\textit{rule-based machine learning (RBML)}} algorithm for classification is FOLD-SE \cite{foldse}. FOLD-SE, being efficient and explainable, generates default rules---a stratified normal logic program---as an \textit{explainable} model from the given input dataset. Both numerical and categorical features are allowed. The generated rules symbolically represent the machine learning model that will predict a label, given a data record. FOLD-SE can also be used for learning rules capturing causal dependencies among features in a dataset. FOLD-SE maintains scalability and explainability, as it learns a relatively small number of rules and literals regardless of dataset size, while retaining good classification accuracy compared to state-of-the-art machine learning methods.

\vspace{-0.2 in}

\section{Overview}
\vspace{-0.07 in}

\subsection{The Problem}
\vspace{-0.05 in}

In high-stakes decision-making systems, individuals (represented as a set of features) often receive undesired negative decisions (e.g., loan denial) from black-box machine learning models. These models lack transparency, making it difficult to understand why a decision was made and what changes are necessary to flip it to a positive outcome. \textit{P2C} automatically identifies these changes.
For example, if John is denied a loan (\textit{initial state $i$}), \textit{P2C} models the set of all (positive) scenarios (\textit{goal set $G$}) where he obtains the loan. Out of these scenarios, John wishes to reach the scenario requiring minimal effort (\textit{minimal causally compliant counterfactual $g\in G$}). The query goal `{\tt ?- reject\_loan(john)}' represents the prediction of the classification model regarding whether John’s loan should be rejected, based on the extracted underlying logic of the model used for loan approval. The (negative) decision in the \textit{initial state $i$} should not apply to any scenario in the \textit{goal set $G$}. The query goal `{\tt ?- reject\_loan(john)}' should be {\tt True} in the initial state \(i\) and {\tt False} for all goals in the \textit{goal set $G$}. The problem is to find a series of interventions, namely, changes to feature values, that will take John from $i$ to $g \in G$.
\vspace{-0.11 in}

\subsection{Solution: \textit{P2C} Approach}\label{sec_P2C_approach}
\vspace{-0.11 in}

P2C finds the path from an initial state $i$ to the \textit{minimal causally compliant counterfactual $g\in G$} with each state represented as feature-value pairs (e.g., credit score: 620; debt: $0$). This is done in a three-step process: \textbf{1) Black-Box Model Approximation:} The black-box model $D$ is approximated using a RBML algorithm (FOLD-SE), generating an explainable surrogate model $H$ that mimics the black-box model’s $D$ decision-making. \textbf{Optional:} \textit{In case the causal model $C$ is not provided, it is learned using FOLD-SE. The learned explainable causal model $C$ is then verified by domain experts to check that indeed causality and not correlation is captured.} \textbf{2) Causal-Aware Counterfactual Search:} Using ASP-based reasoning using $C$ and $D$, P2C identifies causally feasible changes that transition the input from $i$ to $G$. \textbf{Optimized Cost Computation:} Unlike prior methods, P2C distinguishes between direct interventions and automatic causal effects, ensuring cost is assigned only to user-initiated changes (direct changes). Using this, P2C computes the \textit{minimal causally compliant counterfactual $g\in G$}. \textbf{3) Planning:} Using $D$ and $C$, generate a \textit{step-by-step} plan of interventions to go from $i$ to $g\in G$.

This generation of a plan is a \textit{planning problem}. However, unlike the standard \textit{planning problem}, the interventions that take us from one state to another are not mutually independent: there may exist a causal dependency. 
P2C ensures that each intervention respects casual dependencies between variables, offering an explanation of how one action leads to the next. The step-wise approach of \textit{P2C} contrasts with approaches like \textit{MINT}, which applies all interventions simultaneously. Such simultaneous interventions are not helpful for understanding the dynamic changes in systems with complex causal relationships. The objective is to turn a negative decision (\textit{initial state $i$}) into a positive one (\textit{goal state $g$}) through necessary changes to feature values, so that the query goal `{\tt ?- not reject\_loan(john)}' will succeed for $g \in G$.

\textit{P2C} models a path between two scenarios: 1) the negative outcome world (e.g., loan denial, \textit{initial state $i$}), and 2) the positive outcome world (e.g., loan approval, \textit{goal state $g$}) achieved through specific interventions. Both states are defined by specific attribute values (e.g., loan approval requires a credit score $\geq$ 620). \textit{P2C} symbolically computes the necessary interventions to find a path from $i$ to $g$, representing a flipped decision. 
When the decision query (e.g., `{\tt ?- reject\_loan/1}') succeeds (negative outcome), \textit{P2C}
finds the state where this query fails (e.g., `{\tt ?- not reject\_loan/1}' succeeds), which constitutes the \textit{goal state $g$}.
In terms of ASP, the task is as follows: given a world where a query succeeds, compute changes to feature values (accounting for causal dependencies) to reach another world where negation of the query will succeed. Each intermediate world traversed must be viable with respect to the rules, i.e., the traversed worlds must be realistic. We use the s(CASP) query-driven predicate ASP system for this purpose. 

\textit{P2C} employs two kinds of actions: \textbf{1) Direct Actions:} directly changing a feature value, and \textbf{2) Causal Actions:} changing other features to cause the target feature to change, utilizing the causal dependencies between features. These actions guide the individual from \textit{$i$} to \textit{$g$} through intermediate states, suggesting realistic and achievable changes.
Unlike \textit{P2C}, the other approaches \cite{wachter}, \cite{alt_karimi} can output non-viable solutions as they assume feature independence.

\begin{figure*}[htp]
    \centering    \includegraphics[width=12cm]{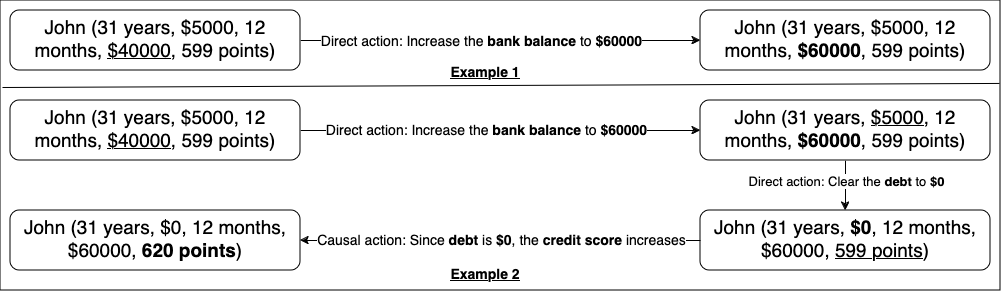}
    \caption{\textbf{Example 1:} John increases his \textit{bank balance} to $\$60000$. \textbf{Example 2:} The \textit{bank balance} and \textit{debt} are directly altered by John. The causal effect of having $\$0$ \textit{debt} increases John's \textit{credit score} to $620\ points$.\vspace{-0.2in}}
    \label{fig_example}
\end{figure*}

\smallskip\noindent\textbf{Example 1: Using direct actions to reach the counterfactual state} \label{eg_1}
Consider a loan application scenario. There are \textbf{five} feature-domain pairs: 1) Age: \{1 year,..., 99 years\}, 2) Debt: \{$\$1$, ..., $\$1000000$\}, 3) Loan Duration: \{$1\ month,\ ...,\ 60\ months$\}, 4) Bank Balance: \{$\$0$, ..., $\$1\ billion$\} and 5) Credit Score: \{$300\ points, ..., 850\ points$\}. John (31 years, $\$5000$, 12 months, $\$40000$, $599\ points$) applies for a loan. Based on the extracted underlying logic of the classifier for loan rejection, the bank denies his loan (negative outcome) as his \textit{bank balance} is \textbf{less than $\$60000$}. To get approval (positive outcome), \textit{P2C} recommends:
\textbf{Initial state}: John (31 years, $\$5000$, 12 months, $ \$40000$, $599\ points$) is denied a loan.
\textbf{Goal state}: John (31 years, $\$5000$, 12 months, $\$60000$, $599\ points$) is approved.
\textbf{Intervention}: Increase the \textit{bank balance} to $\$60000$. As shown in Fig. \ref{fig_example}, the direct action flips the decision, making John eligible for the loan.


\vspace{-0.1in}
\vspace{-0.1in}
\smallskip\noindent\textbf{Example 2: Utility of Causal Actions}\\
The extracted underlying logic of the classifier for loan rejection produces two rejection rules: 1) individuals with a \textit{bank balance} of \textbf{less than $\$60000$}, and 2) individuals with a \textit{credit score} below $620$. John (31 years, $\$5000$, $12$ months, $ \$40000$, $599\ points$) is denied a loan (negative outcome) but wants approval (positive outcome). Without causal knowledge, the solution would be: \textbf{Interventions}: 1) Change the \textit{bank balance} to $\$60000$, and 2) the \textit{credit score} to $620\ points$. However, \textit{credit score} cannot be changed directly. To realistically increase the \textit{credit score}, the bank’s guidelines suggest \textbf{having no debt}, indicating a causal dependency between \textit{debt} and \textit{credit score}. \textit{P2C} recommends: \textbf{Initial state}: John (31 years, $\$5000$, 12 months, $ \$40000$, $599\ points$) is denied a loan. \textbf{Goal state}: John (31 years, $\$0$, 12 months, $\$60000$, $620\ points$) is approved for a loan. \textbf{Interventions}: 1) John increases \textit{bank balance} to $\$60000$, and 2) \textit{reduces debt} to \$0 to increase his \textit{credit score}. From Figure \ref{fig_example} clearing the \textit{debt} (direct action) leads to an increase in \textit{credit score} (causal action), making John eligible for the loan. Intermediate states (e.g., `$\$5000$ in \textit{debt}' and `$\$0$ in \textit{debt}') represent the path to $g$. We demonstrated how leveraging causal dependencies between features leads to realistic desired outcomes through appropriate interventions.


The challenge now is (i) identifying causal dependencies either by applying common knowledge or through the help of domain experts to the rules learned by using rule-based machine learning algorithms \textit{(RBML)} \textbf{(in case the causal model is not provided to us)}, and (ii) computing the sequence of necessary interventions while avoiding repeating states--- a known issue in planning. \textit{P2C} addresses these by generating the path from $i$ to $g\in G$.
\vspace{-0.15 in}
\section{Methodology}\label{sec_methodology}
\vspace{-0.1 in}

We next outline the methodology used by \textit{P2C} to generate paths from the initial state (\textit{negative outcome}) to the goal state (\textit{positive outcome}). In \textbf{Step 1}, we utilize the causal and decision rules to obtain a minimal causally compliant counterfactual $g\in G$. For \textbf{Step 2}, unlike traditional planning problems where actions are typically independent, our approach involves \textit{interdependent} actions governed by causal rules $C$. This ensures that the effect of one action can influence subsequent actions, making interventions realistic and causally consistent. 
Note that in cases where the causal model is not provided, the \textit{P2C} framework uses the FOLD-SE \textit{RBML} algorithm to automatically compute causal dependency rules. These rules have to be either verified by a human, or commonsense knowledge must be used to verify them automatically. This is important, as \textit{RBML} algorithms can identify a correlation as a causal dependency. \textit{P2C} uses the former approach. We next define specific terms.

\begin{definition}[\textbf{State Space (S)}]\label{defn:S}
$S$ represents all combinations of feature values. For domains \( D_1,..., D_n \) of the features \( F_1,..., F_n\), $S$ is a set of possible states $s$, where each state is defined as a tuple of feature values \( V_1, ..., V_n \). 

\vspace{-0.2in}
\begin{align*}
s\in S\ where\ S = \{ (V_1,V_2,...,V_n)\ |\  
V_i \in D_i,\ for\ each\ i\ in\ 1,...,n \}     
\end{align*}
\end{definition}

\noindent \textit{E.g., an individual John: \textit{s = $(31$ years, $\$5000$, $12$, $\$40000$, $599$ points$)$}, where\ $s\in S$.}
\vspace{-0.05 in}
\begin{definition}[\textbf{Causally Consistent State Space ($S_C$)}]\label{defn:S_C}
$S_C$ is a subset of $S$ where all causal rules are satisfied. $C$ represents a set of causal rules over the features within a state space $S$. Then, $\theta_C: P(S) \rightarrow P(S)$ (where $P(S)$ is the power set of S) is a function that defines the subset of a given state sub-space $S'\subseteq S$ that satisfy all causal rules in C.

\vspace{-0.1 in}
\begin{equation*}
\theta_C(S') = \{ s \in S' \mid s\ satisfies\ all\ causal\ rules\ in\ C\}
\end{equation*}
\begin{equation*}
S_C = \theta_C(S) 
\end{equation*}


\noindent 
E.g., causal rules state that if $debt$ is $0$, the credit score should be above 599, then instance $s_1$ = (31\ years,\ \$0,\ 12,\ \$40000,\ 620\ points) is causally consistent, and instance $s_2$ = (31\ years,\ \$0,\ 12,\ \$40000,\ 400\ points) is causally inconsistent. 

\end{definition}

\noindent 
In a traditional planning problem, allowed actions in a given state are independent, i.e., the result of one action does not influence another. In \textit{P2C}, causal actions are interdependent, governed by $C$. 

\vspace{-0.05 in}
\begin{definition}[\textbf{Decision Consistent State Space ($S_Q$)}]\label{defn:S_Q}
%
$S_Q$ is a subset of $S_C$ where all decision rules are satisfied. $Q$ represents a set of rules that compute some external decision for a given state. $\theta_Q: P(S) \rightarrow P(S)$ is a function that defines the subset of the causally consistent state space $S'\subseteq S_C$ that is also consistent with decision rules in $Q$:
\begin{equation*}
\theta_Q(S') = \{ s \in S' \mid s\ satisfies\ any\ decision\ rule\ in\ Q\} \label{defn:3_1}
\end{equation*}

Given $S_C$ and $\theta_Q$, we define the decision consistent state space $S_Q$ as 
\vspace{-0.055 in}
\begin{equation*}
S_Q = \theta_Q(S_C) = \theta_Q(\theta_C(S)) 
\end{equation*}
\vspace{-0.03 in}
\noindent E.g., an individual John whose loan has been rejected: \textit{s = ($31$ years, $\$0$, $12$, $\$40000$, $620$ points)}, where\ $s\in S_Q$.

\end{definition}
\vspace{-0.1 in}
\begin{definition}[\textbf{Initial State ($i$)}]\label{defn:I}
$i$ is the starting point with an undesired outcome. Initial state $i$ is an element of the \textit{causally consistent state space} $S_C$
\vspace{-0.1 in}
\begin{equation*}
i \in S_C 
\end{equation*}

\noindent For example, $i = (31\ years,\ \$0,\ \$40000,\ 620\ points)$
\end{definition}
\vspace{-0.1 in}
\begin{definition}[\textbf{Actions}]\label{defn:action}
The set of actions $A$ includes all possible interventions (actions) that can transition a state from one to another within the state space. Each action $a\in A$ is defined as a function that maps $s$ to a new state $s'$. 
\vspace{-0.1 in}
\begin{equation*}
    a: S\rightarrow S\ \mid where\ a\in A
\end{equation*}
\vspace{-0.25in}

Actions are divided into: 1) Direct Actions: Directly change the value of a single feature of a state $s$, e.g.,  Increase bank balance from $\$40000$ to $\$60000$. 2) Causal Actions: Change the value of a target feature by altering related features, based on causal dependencies. It results in a causally consistent state with respect to C, e.g., reduce debt to increase the credit score. 
\end{definition}
\vspace{-0.1 in}
\begin{definition}[\textbf{Transition Function}]\label{defn:delta}
A transition function $\delta: S_C \times A \rightarrow S_C$ maps a causally consistent state to the set of allowable causally consistent states that can be reached in a single step, and is defined as: 
\vspace{-0.05in}
  \[\delta(s,a) = \left\{\begin{array}{l}
     a(s) \textit{ if } a(s) \in S_C\\
     \delta(a(s),a') \textit{ with } a\in A, a'\in A \textit{, otherwise}
  \end{array}\right.\]

\noindent $\delta$ models a function that repeatedly takes actions until a causally consistent state is reached. In \textbf{example 1}, $\delta$ suggests  changing the \textit{bank balance} from $\$40000$ to $ \$60000$: 
$\delta(31\ years,\$5000,\$40000,599) = $ $(31\ years,\$5000,\$60000,599)$


\end{definition}
\vspace{-0.1 in}
\begin{definition}[\textbf{Counterfactual Generation (CFG) Problem}]\label{problem_statement}
A counterfactual generation (CFG) problem is a 4-tuple $(S_C,S_Q,I,\delta)$ where $S_C$ is causally consistent state space, $S_Q$ is the decision consistent state space, $I\in S_C$ is the initial state, and $\delta$ is a transition function.
\end{definition}
\vspace{-0.22 in}
\begin{definition}[\textbf{Goal Set}]\label{defn:G}
The goal set $G$ is the set of desired outcomes that do not satisfy the decision rules $Q$. For the Counterfactual Generation (CFG) problem $(S_C,S_Q,I,\delta)$: $G = \{ s \in S_C| s\not\in S_Q\}$. 

\noindent $G$ includes all states in $S_C$ that do not satisfy $S_Q$. 
For \textbf{example 1}, \textit{g = (31 years,\ \$0,\ \$60000,\ 620\ points) $\mid$ g$\in$G.}
\end{definition}
\vspace{-0.2 in}
\begin{definition}[\textbf{Solution Path}]\label{solution_path}
A solution to the problem $(S_C,S_Q,I,\delta)$ with goal set $G$ is a path:
\vspace{-0.15 in}
\begin{align*}
    &s_0, s_1, \dots, s_m \ \text{where} \ s_j \in S_C \ \text{for all} \ j \in \{0, \dots, m\}, \\
    &such\ that\ s_0 = I; s_m = G; s_0,...,s_{m-1} \not\in G; s_{i+1} \in \delta(s_i) \ \text{for} \ i \in \{0, \dots, m-1\}
\end{align*}
For \textbf{example 1}, individuals with less than $\$60000$ in their account are ineligible for a loan, thus the state of an ineligible individual $s\in S_Q$ might be $s = (31\ years,\$5000,12,\$40000,599\ points)$. The goal set has only one goal state $g\in G$ given by $g = (31\ years,\$5000,12,\$60000,599\ points)$. The path from $s$ to $g$ is \{\textit{(31\ years, \$5000,12, \$40000, 599\ points)}$\rightarrow Direct Action:\ Increase\ bank\ balance$ $\rightarrow $\textit{(31\ years,\$5000,12,\$60000,599\ points)}$\}$. Here, the path has only 2 states as only changing the \textit{bank balance} to be $\$60000$ is needed to reach the goal state.
\vspace{-0.15 in}
\end{definition}

\smallskip\noindent\textbf{Algorithms:} 
We now detail the algorithmic framework of P2C, which produces the \textit{minimal causally compliant counterfactual (MCCC)} and constructs feasible paths from the original instance $i$ to the \textit{MCCC}. The workflow comprises three stages: 1) Rule extraction: derive a rule-based approximation $Q$ of the black-box classifier $M$ using a rule-based machine-learning (RBML) algorithm; 2) MCCC Search: Identify the \textit{MCCC} $s^*$ that satisfies the causal model $C$; and 3) Path finding: compute a causally compliant sequence of actions that transforms the initial instance $i$ into $s^*$. The remainder of this section presents the top-level procedure, followed by detailed pseudocode for every subroutine.

\vspace{-0.2 in}
\subsection{Algorithm for P2C:}
\vspace{-0.05 in}
\textbf{Algorithm \ref{p2c_algo2}} provides the pseudocode for obtaining a feasible path from the original instance $i$ to a \textit{minimal causally compliant counterfactual (MCCC)} through the following three-step procedure:
\textbf{1) Extract Logic}: Algorithm \ref{alg_underlying_logic} extracts a rule-based approximation $Q$ of the black-box classifier $M$ using an RBML algorithm $R$. A causal model $C$ is supplied; if absent, we learn a candidate $C$ from the data and have domain experts validate its causal soundness.
\textbf{2) MCCC search}: Given $Q$, $C$, the initial state $i$, feature-change weights $W$, and state-space $S$ (Sec. \ref{sec_methodology} Def. \ref{defn:S}), Algorithm \ref{alg_obtain_cf} returns the MCCC $s^*$.
\textbf{3) Find Path}: Algorithm \ref{alg_path} then computes a feasible path from $i$ to $s^*$.

\vspace{-0.15in}
\subsection{Algorithm to Extract Decision Rules}
\vspace{-0.05in}

The function `\textit{\textbf{extract\_logic}}' extracts the underlying logic of the classification model used for decision-making. 
\textbf{Algorithm \ref{alg_underlying_logic}} provides the pseudocode for `\textit{\textbf{extract\_logic}}', which takes the original classification model $M$, input data $H$, and a \textit{RBML} algorithm $R$ as inputs and returns $Q$, the underlying logic of the classification model.
If $M$ is already rule-based, we set $Q=M$. Otherwise, we first label the data $H$ with the model $M$ and set the label to $V\ =\ predict(M(H))$. We then train the \textit{RBML} algorithm $R$ on $H$ and $V$ to return a surrogate rule set $Q$. This $Q$ captures the decision logic responsible for the undesired outcomes.

\vspace{-0.2in}

\subsection{Algorithm to Obtain the Minimal Effort Counterfactual}

\vspace{-0.1in}
\textbf{Algorithm \ref{alg_obtain_cf}} provides the pseudo-code for `\textit{\textbf{min\_cg}}' which returns the \textit{MCCC} given the initial state $i$, Decision Rules $Q$, Causal Model $C$ and State Space $S$ (Sec. \ref{sec_methodology} Def. \ref{defn:S}). It calls \textbf{is\_counterfactual} which is {\tt True} if the state is a causally consistent counterfactual.
For each candidate state, \textit{min\_cf} calls \textit{is\_counterfactual} and sets zero to the weights of features changed automatically by causal dependencies. Hence causally compliant incur \textbf{zero} cost. Finally, \textit{compute\_weighted\_Lp} measures the distance from the initial state to each valid counterfactual, taking into account only direct user-initiated changes. The algorithm then returns the counterfactual with the lowest total cost.
\vspace{-0.25 in}
\subsection{Algorithm to Obtain a path to the Counterfactual} \label{sec_alg_path}
\vspace{-0.1 in}
We next describe our algorithm to find the goal states and compute the solution paths, \textit{\textbf{find\_path}}, which makes use of the following functions: (i) \textbf{not\_member}: checks if an element is: \textit{a}) \textbf{not} a member of a list, and \textit{b}) Given a list of tuples, \textbf{not} a member of any tuple in the list. (ii) \textbf{drop\_inconsistent}: given a list of states [$s_0,...,s_k$] and a set of Causal rules $C$, it drops all the inconsistent states resulting in a list of consistent states with respect to $C$. (iii) \textbf{get\_last}: returns the last member of a list. (iv) \textbf{pop}: returns the last member of a list. (v) \textbf{intervene}: performs interventions/ makes changes to the current state through a series of actions and returns a list of visited states. The interventions are causally compliant. Further details are available in the supplement. 

\noindent Function `\textit{\textbf{find\_path}}' implements the Solution Path $P$ of Definition \ref{solution_path}. Its purpose is to find a path to the counterfactual state. \textbf{Algorithm \ref{alg_path}} provides the pseudo-code for `\textit{\textbf{find\_path}}', which takes as input an Initial State $i$, a set of Causal Rules $C$, Decision Rules $Q$, and Actions $A$. It returns a path to the counterfactual state/goal state $g\in G$ for the given $i$ as a list `\textit{visited\_states}'. Unrealistic states are removed from `\textit{visited\_states}' to obtain a `\textit{candidate\_path}'.

Initially, $s=i$. The function checks if the current state $s$ is a counterfactual. If $s$ is already a counterfactual, `\textit{\textbf{find\_path}}' returns a list containing $s$. If not, the algorithm moves from $s=i$ to a new causally consistent state $s'$ using the `\textit{\textbf{intervene}}' function, updating `\textit{visited\_states}' with $s'$. It then checks if $s'$ = $s^*$, i.e., the (\textit{MCCC}). If {\tt True}, the algorithm drops all inconsistent states from `\textit{visited\_states}' and returns the `\textit{candidate\_path}' as the path from $i$ to $s'$. If not, it updates `$current\_state$' to $s'$ and repeats until reaching the  \textit{minimal causally compliant counterfactual (MCCC)} state $s^*$. The algorithm ends when the last state in the list `\textit{visited\_states}' is $s^*$, i.e. $g\in G$.


\vspace{-0.2in}

\subsubsection{Discussion:} \textbf{(i)} Certain features are immutable or restricted: for example, \textit{age} cannot decrease and \textit{credit score} cannot be altered directly. To restrict the set of admissible actions without adding new states, we introduce plausibility constraints. These constraints are encoded in the action set in Algorithms \ref{alg_path}.   
\textbf{(ii)} \textit{Direct-Path Length (DPL)}:  \textit{P2C} begins with \textit{DPL} = 1, searching for counterfactuals achievable via a single direct change, i.e., change through a single direct action; if none exist, it incrementally increases \textit{DPL} until a solution is found. This guarantees counterfactuals that are minimal and causally consistent.
\textbf{(iii)} Since \textit{P2C} relies heavily on backtracking, we implement it in the goal-directed ASP system s(CASP), which provides built-in backtracking support.

\vspace{-0.1 in}
\begin{algorithm}[!h]
\caption{\textsc{P2C}: Path to a Causal Counterfactual}\label{p2c_algo2}
\begin{algorithmic}[1]
  \REQUIRE Classifier $M$, dataset $H$, RBML learner $R$, Initial state $i$, Weights $W$ 
  \STATE $Q = \textit{extract\_logic}(M, H, R)$           // Decision rules
  \STATE $s^{*} = \textit{min\_cf}(i, S, W, C, Q)$  // Minimal causally compliant CF
  \STATE \textit{candidate\_path} $= \textit{find\_path}(i, S, W, C, Q, s^{*})$ // Plan from $i$ to $s^{*}$
  \STATE RETURN \textit{candidate\_path}
\end{algorithmic}
\end{algorithm}
\vspace{-0.3 in}


\vspace{-0.2in}
\begin{algorithm}[!h]
\caption{\textbf{extract\_logic}: Extract the underlying logic of the classification model}
\label{alg_underlying_logic}
\begin{algorithmic}[1]
    \REQUIRE Original Classification model $M$, Data $H$, \textit{RBML} Algorithm \textit{R}:
    \IF{ $M$ is rule-based}
    \STATE Set \textit{Q} = \textit{M} // Decision Rules are the rules of model \textit{M}
    \ELSE 
    \STATE Set \textit{V} = \textbf{predict(}\textit{M(H)}\textbf{)} // For input data \textit{H}, predict the labels as \textit{V}
    \STATE Set \textit{Q} = \textbf{train(}\textit{R(H,V)}\textbf{)} // Train \textit{R} on \textit{H} and \textit{V} to obtain Decision Rules \textit{Q}
    \ENDIF
    \STATE Return \textit{Q}

\end{algorithmic}
\end{algorithm}

\vspace{-0.5in}
\begin{algorithm}[!h]
\caption{\textbf{min\_cf}: Find the Minimal CF}
\label{alg_obtain_cf}
\begin{algorithmic}[1]
    \REQUIRE Initial state $i$, States $S$, Causal Model \textit{C}, Decision Rules \textit{Q}, Weights \textit{W}
    \STATE bestCost = $\infty$ // Initialize best cost to a large value
    \STATE $s^*$ = \textbf{NULL} // Optimal counterfactual state not found yet
    \FORALL{$s \in S$}
    \STATE //Check if $s$ is a counterfactual and adjust weights
    \STATE $(\textit{isValid},\ \textit{adjWeights}) = \textbf{is\_counterfactual}(s,\ C,\ Q,\ W)$ 
    \IF{\textit{isValid} = TRUE}
    \STATE $cost$ = \textbf{compute\_weighted\_Lp}($i$, $s$,\ \textit{adjWeights}, $p$)
    \IF{\textit{cost} $<$ $bestCost$}
    \STATE $bestCost$ = $cost$; $s^*$ = $s$
    \ENDIF
    \ENDIF
    \ENDFOR
    
    \STATE Return $(s^*,\ bestCost)$

\end{algorithmic}
\end{algorithm}
\vspace{-0.2in}

\begin{algorithm}[!ht]
\caption{\textbf{find\_path}: Obtain a path to the counterfactual state}
\label{alg_path}
\begin{algorithmic}[1]
    \REQUIRE Initial State $i$, States $S$, Feature Weights $W$, Causal Rules $C$, Decision Rules $Q$, Minimum Counterfactual $s^*$, Actions $a\in A$:

    \STATE\label{PS21} Create an empty list \textit{visited\_states} that tracks the list of states traversed (so that we avoid revisiting them).
    \STATE Append ($i$, [~]) to \textit{visited\_states} 
    \WHILE{\textit{get\_last(visited\_states)\ $\neq$ $s^*$ }}
    \STATE Set \textit{visited\_states=intervene(visited\_states,C,A)}
    \ENDWHILE
    \STATE \textit{candidate\_path = drop\_inconsistent(visited\_states)}
    \STATE Return \textit{candidate\_path}

\end{algorithmic}
\end{algorithm}

\subsection{Soundness}
\vspace{-0.05 in}

\begin{definition}[CFG Implementation]\label{defn:impl_cfg}
When Algorithm \ref{alg_path} is executed with the inputs: Initial State $i$ (Definition \ref{defn:I}), States Space $S$ (Definition \ref{defn:S}), Set of Causal Rules $C$ (Definition \ref{defn:S_C}), Set of Decision Rules $Q$ (Definition \ref{defn:S_Q}), and Set of Actions $A$ (Definition \ref{defn:action}), a CFG problem $(S_C,S_Q,I,\delta)$ (Definition \ref{problem_statement}) with causally consistent state space $S_C$ (Definition \ref{defn:S_C}), Decision consistent state space $S_Q$ (Definition \ref{defn:S_Q}), Initial State $i$ (Definition \ref{defn:I}), the transition function $\delta$ (Definition \ref{defn:delta}) is constructed. 
    
\end{definition}



\vspace{-0.1in}
\begin{definition}[\textit{Candidate path}]\label{defn:candidate_path}
Given the counterfactual $(S_C,S_Q,I,\delta)$ constructed from a run of  algorithm \ref{alg_path}, the return value (candidate path) is the resultant list obtained from removing all elements containing states $s'\not\in S_C$.     
\end{definition}

\noindent Definition \ref{defn:impl_cfg} maps the input of Algorithm \ref{alg_path} to a \textit{CFG problem} (Definition \ref{problem_statement}). \textit{Candidate path} maps the result of Algorithm \ref{alg_path} to a possible solution (Definition \ref{solution_path}) of the corresponding CGF problem. From Theorem 1 \textit{(proof in supplement)}, the \textit{candidate path} (Definition \ref{defn:candidate_path}) is a solution to the corresponding \textit{CFG problem} implementation (Definition \ref{defn:impl_cfg}).

\noindent \textit{Theorem 1} 
Soundness: Given a CFG $\mathbb{X}$=$(S_C,S_Q,I,\delta)$, constructed from a run of Algorithm \ref{alg_path} \& a corresponding candidate path $P$, $P$ is a solution path for $\mathbb{X}$. Proof is provided in the supplement 
.

\vspace{-0.2in}

\section{Experiments}\label{sec_experiments}
\vspace{-0.1 in}
Our experiments address three questions: 1) How effective is P2C's refined cost metric; 2) How does P2C scale as the search space increases?; and 3) Do the path-to-counterfactuals produced by P2C outperform those produced by a standard path finder? 
We evaluate on: \textit{Adult} \cite{adult}, \textit{Statlog (German Credit)} \cite{german}, and the \textit{Car evaluation} \cite{car} datasets. These datasets include demographic and decision labels such as credit risk (`\textit{good}' or `\textit{bad}'), income (‘$=<\$50k/year$’ or ‘$>\$50k/year$’), and used car acceptability. We relabeled the car evaluation dataset -\textit{`acceptable'} or \textit{`unacceptable'}- to generate the counterfactuals. For the German dataset, \textit{P2C} identifies paths that convert a `\textit{bad}' credit rating to `\textit{good}' to determine the criteria for a favourable credit risk. Similarly, \textit{P2C} identifies paths for converting the undesired outcomes in the \textit{Adult} and \textit{Car Evaluation} datasets-‘$=<\$50k/year$’ and \textit{`unacceptable'}-to their counterfactuals: ‘$>\$50k/year$’ and \textit{`acceptable'}. We use \textit{P2C} to obtain counterfactual paths. Further details and the implementation are provided in the supplement.

\vspace{-0.2in}
\subsection{Comparison of Counterfactual Proximity}
\vspace{-0.1 in}
\noindent To show the effectiveness of P2C's refined cost metric, we compare it against another ASP based causally compliant counterfactual method: C3G \cite{C3G}. C3G and P2C use the same Decision and Causal Rules. While C3G is not designed to produce L1 and L2 norms (as it returns counterfactual ranges), we have modified it to obtain each individual counterfactual point and hence obtain L1 and L2 norms.
Table \ref{tbl:c3g_mc3g} demonstrates that for datasets with causal dependencies, P2C consistently produces counterfactuals closer or equal to that of C3G across all metrics---nearest, furthest, and average distances---regardless of norm---L0 or L1 or L2---used. This is inspite of C3G and P2C using the same decision rules $Q$ and the same causal rules $C$. This is because P2C correctly accounts for causal dependencies, treating causally induced changes as cost-free, whereas C3G incorrectly assigns a cost to all feature modifications, including those that occur naturally (causal effect).  For both the Adult and German datasets, P2C counterfactuals exhibit lower nearest, furthest, and average distances/cost than C3G. This highlights that P2C identifies more efficient intervention strategies, ensuring that users receive recourse recommendations requiring minimal effort while remaining causally compliant. However for the Cars dataset, since we have not identified any causal dependencies, the performance of P2C is identical to C3G. Overall for causal datasets, the reduced distance across metrics confirms that P2C outperforms C3G in generating counterfactuals with minimum cost.

\vspace{-0.2 in}
\subsection{Reducing Search Space for Scalability}
\vspace{-0.3in}
\begin{table}[h!]
\centering
\begin{tabular}{|p{1.5cm}|p{2cm}|p{2cm}|p{3cm}|p{2cm}|}
\cline{1-5}
\textbf{Dataset} & \textbf{Search Space Size} & \textbf{Avg. Time to cf (ms)} & \textbf{Reduced Search Space Size} & \textbf{Avg. Time to cf (ms)} \\
\cline{1-5}
Adult  & 11,225,088 & 185.53 & 6,414,336 & 107.57 \\
\cline{1-5}
German & 13,200     & 262.34 & 2,640     & 74.11  \\
\cline{1-5}
Cars & 1728     & 1.963 & 432     & 1.878  \\
\cline{1-5}
\end{tabular}
\caption{Time to generate a counterfactual reduces as Search Space Size decreases}
\label{tab:scalability}
\end{table}
\vspace{-0.2in}


\noindent \textit{P2C} can be computationally heavy as it explores an expanding search space for counterfactuals. For scalability, we consolidated feature values independent of decision/causal rules into placeholders. For example, the rule: {\tt loan\_accept:-} {\tt marital\_status(married).} with  domain: \textit{{married,unmarried,separated}}, we replace \textit{unmarried, separated} with a placeholder (e.g., \textit{ph\_married}), reducing the domain size from 3 to 2. This reduces the search-space (the product of effective domain sizes) without altering the core algorithm. 
From Table \ref{tab:scalability}, we see that place-holding not only reduces the search-space but also reduces the average time to find a counterfactual. 

\vspace{-0.2in}

\subsection{P2C: Comparison of the quality of the Path Generated}
\vspace{-0.07in}

We evaluate P2C against counterfactual-based approaches: Borderline Counterfactuals \cite{wachter}, DiCE \cite{DiCE}, MACE \cite{alt_karimi}, MINT \cite{ref_4_karimi_2} and C3G \cite{C3G}. Since none of these methods natively support path-finding, we use a standard planner to find a path to the counterfactual state. We aim to show that the path-finding approach of P2C is causally compliant and does not produce any illegal actions (such as directly increasing the credit score) compared to using a standard path-finding algorithm. We use the following metrics:  \textbf{1) Causal Compliance:} $TRUE$/$FALSE$ values indicate whether the method is designed to handle causal dependencies;  \textbf{2) Causal Consistency:} The percentage of the generated counterfactuals that are causally compliant.; \textbf{3) Path to counterfactuals:} $TRUE$/$FALSE$ values indicate whether the path to the counterfactual contains legal actions, i.e. actions that respect causal dependencies. 

As seen in Table  \ref{tbl:combined_adult_german}, P2C produces causally compliant counterfactuals 100\% of the time. Unlike a similar causally compliant counterfactual method such as MINT and C3G, it provides a path with only legal actions leading to the counterfactual state. This path is designed to utilize causal dependencies and generate causal actions, something the standard path-finding algorithm cannot do. The only case where the standard path-finder matches the performance of P2C is for the Cars dataset when we do not know the causal dependencies.


\begin{table}[ht]
    \centering
    \renewcommand{\arraystretch}{1.3} 
    \begin{tabular}{|p{1.1cm}|p{0.8cm}|p{1.0cm}|p{1.1cm}|p{0.9cm}|p{1.0cm}|p{1.1cm}|p{1.0cm}|p{1.0cm}|p{1.1cm}|p{0.6cm}|}
        \cline{1-11}
        \multirow{4}{*}{\makecell{Dataset}} & \multirow{4}{*}{Model}
            & \multicolumn{9}{c|}{\textbf{K = 20}} \\ \cline{3-11}
        & & \multicolumn{9}{c|}{\textbf{Metric Used to Sort the Closest}} \\ \cline{3-11}
        & & \multicolumn{3}{c|}{\textbf{L1}}
            & \multicolumn{3}{c|}{\textbf{L2}}
            & \multicolumn{3}{c|}{\textbf{L0}} \\ \cline{3-11}
        & & Nearest & Furthest & Avg.
            & Nearest & Furthest & Avg.
            & Nearest & Furthest & Avg.
            \\ \cline{1-11}
        \multirow{2}{*}{Adult}  & C3G & 1.012 & 1.470 & 1.296 & 1.012 & 1.221 & 1.113 & 1 & 1 & 1 \\ \cline{2-11}
                                & P2C & 0.701 & 1.031 & 0.907 & 0.701 & 0.903 & 0.829 & 1 & 1 & 1 \\ \cline{1-11}
        \multirow{2}{*}{German} & C3G & 3.3342 & 3.3415 & 3.3378 & 1.76399 & 1.76401 & 1.7640 & 4 & 4 & 4 \\ \cline{2-11}
                                & P2C & 2.3342 & 2.3415 & 2.3379 & 1.45316 & 1.45318 & 1.45317 & 3 & 3 & 3 \\ \cline{1-11}
        \multirow{2}{*}{Cars}   & C3G & 1 & 3 & 2.3 & 1 & 1.732 & 1.499 & 1 & 3 & 2.3 \\ \cline{2-11}
                                & P2C & 1 & 3 & 2.3 & 1 & 1.732 & 1.499 & 1 & 3 & 2.3 \\ \cline{1-11}
    \end{tabular}
    \caption{Comparison of Nearest and Furthest Counterfactuals for C3G and P2C}
    \label{tbl:c3g_mc3g}
\end{table}
\vspace{-0.5in}

\begin{table}[h!]
\centering
\begin{tabular}{lcccc}
\cline{1-5}
\textbf{Dataset} & \textbf{Model} & \makecell{Causally\\ Compliant} & \makecell{Cf Causal\\ Consistency (\%)} & \makecell{Path contains \\ only legal actions} \\ 
\cline{1-5}
\multirow{5}{*}{\textbf{Adult}}
  & Borderline-CF & FALSE      & 30  & FALSE \\
  & DiCE          & Indirectly & 80  & FALSE \\
  & MACE          & FALSE      & 80  & FALSE \\
  & MINT          & TRUE       & 100 & FALSE \\
  & C3G          & TRUE       & 100 & FALSE \\
  & \textbf{P2C}          & \textbf{TRUE}       & \textbf{100} & \textbf{TRUE}  \\
\cline{1-5}
\multirow{5}{*}{\textbf{German}}
  & Borderline-CF & FALSE      & 80  & FALSE \\
  & DiCE          & Indirectly & 80  & FALSE \\
  & MACE          & FALSE      & 20  & FALSE \\
  & MINT          & TRUE       & 100 & FALSE \\
  & C3G          & TRUE       & 100 & FALSE \\
  & \textbf{P2C}          & \textbf{TRUE}       & \textbf{100} & \textbf{TRUE}  \\
\cline{1-5}
\multirow{5}{*}{\textbf{Car}}
  & Borderline-CF & FALSE      & N/A  & TRUE \\
  & DiCE          & Indirectly & N/A  & TRUE \\
  & MACE          & FALSE      & N/A  & TRUE \\
  & MINT          & TRUE       & N/A & TRUE \\
  & C3G          & TRUE       & N/A & TRUE \\
  & \textbf{P2C}          & \textbf{TRUE}       & N/A & \textbf{TRUE}  \\
\cline{1-5}
\end{tabular}
\caption{Performance of P2C against counterfactual based methods}
\label{tbl:combined_adult_german}
\end{table}
\vspace{-0.6in}

\section{Conclusion and Future Work}
\vspace{-0.1in}

The main contribution of this paper is the \textit{P2C} framework, which automatically generates paths to a minimal causally compliant counterfactual for any machine learning model---statistical or rule-based. \textit{P2C} is model agnostic. By incorporating Answer Set Programming (ASP), \textit{P2C} ensures that the counterfactual generated is causally compliant and delivered through a sequence of actionable interventions, making it more practical for real-world applications compared to current counterfactual methods. While a limitation of \textit{P2C} lies in it being computationally expensive, a proposed solution involves mapping multiple independent feature values to a single placeholder value to reduce the search space and in turn reduce the complexity. Additionally, \textit{P2C} is limited to tabular data, future tasks will explore extending \textit{P2C} to non-tabular data-- such as image classification tasks \cite{parth_nesy}.
%
%
%
\bibliographystyle{splncs04}
\bibliography{mybibliography}

\begin{thebibliography}{10}
\providecommand{\url}[1]{\texttt{#1}}
\providecommand{\urlprefix}{URL }
\providecommand{\doi}[1]{https://doi.org/#1}

\bibitem{scasp-iclp2018}
Arias, J., Carro, M., Salazar, E., Marple, K., Gupta, G.: Constraint answer set programming without grounding (2018). \doi{10.1017/S1471068418000285}

\bibitem{baral}
Baral, C.: Knowledge representation, reasoning and declarative problem solving. Cambridge University Press (2003)

\bibitem{adult}
Becker, B., Kohavi, R.: {Adult}. UCI Machine Learning Repository (1996), {DOI}: https://doi.org/10.24432/C5XW20

\bibitem{ref_asp_cf}
Bertossi, L.E., Reyes, G.: Answer-set programs for reasoning about counterfactual interventions and responsibility scores for classification. In: Proc. ILP. LNCS (2021)

\bibitem{car}
Bohanec, M.: {Car Evaluation}. UCI Machine Learning Repository (1997), {DOI}: https://doi.org/10.24432/C5JP48

\bibitem{cacm-asp}
Brewka, G., Eiter, T., Truszczynski, M.: Answer set programming at a glance (2011)

\bibitem{C3G}
Dasgupta, S., Shakerin, F., Arias, J., Salazar, E., Gupta, G.: {C3G:} causally constrained counterfactual generation. In: PADL. LNCS (2025)

\bibitem{gelfond-kahl}
Gelfond, M., Kahl, Y.: Knowledge representation, reasoning, and the design of intelligent agents: Answer Set Programming approach. Cambridge Univ. Press (2014)

\bibitem{german}
Hofmann, H.: {Statlog (German Credit Data)}. UCI Machine Learning Repository (1994), {DOI}: https://doi.org/10.24432/C5NC77

\bibitem{alt_karimi}
Karimi, A., Barthe, G., Balle, B., Valera, I.: Model-agnostic counterfactual explanations for consequential decisions. In: AISTATS. PMLR (2020)

\bibitem{ref_4_karimi_2}
Karimi, A., Sch{\"{o}}lkopf, B., Valera, I.: Algorithmic recourse: from counterfactual explanations to interventions. In: Proc. ACM FAccT. pp. 353--362 (2021)

\bibitem{DiCE}
Mothilal, R.K., Sharma, A., Tan, C.: Explaining machine learning classifiers through diverse counterfactual explanations. In: FAT* '20:. {ACM} (2020)

\bibitem{parth_nesy}
Padalkar, P., Wang, H., Gupta, G.: Nesyfold: {A} framework for interpretable image classification. In: Proc. AAAI. pp. 4378--4387. {AAAI} Press (2024)

\bibitem{SCM}
Pearl, J.: {Causal inference in statistics: An overview} (2009)

\bibitem{ref_mace_2}
Russell, C.: Efficient search for diverse coherent explanations. In: Proc. ACM FAT (2019)

\bibitem{ref_github}
Sopam: Supplementary (2025), \url{https://anonymous.4open.science/r/Logic-DBD7/}

\bibitem{ref_mace_1}
Tolomei, G., Silvestri, F., Haines, A., Lalmas, M.: Interpretable predictions of tree-based ensembles via actionable feature tweaking. In: Proc. ACM SIGKDD (2017)

\bibitem{ref_2_ustun}
Ustun, B., Spangher, A., Liu, Y.: Actionable recourse in linear classification. In: Proc. FAT. pp. 10--19 (2019)

\bibitem{wachter}
Wachter, S., Mittelstadt, B.D., Russell, C.: Counterfactual explanations without opening the black box: Automated decisions and the {GDPR} (2017), \url{http://arxiv.org/abs/1711.00399}

\bibitem{foldse}
Wang, H., Gupta, G.: {FOLD-SE:} an efficient rule-based machine learning algorithm with scalable explainability  \textbf{PADL'24, Springer LNCS 14512},  37--53 (2024)

\end{thebibliography}

\newpage
\section{Supplementary Material}
Link to the code can be found in the reference \cite{ref_github}, i.e., \\ https://anonymous.4open.science/r/Logic-DBD7/
\section{Algorithms}

\subsection{Finding the Minimum Counterfactual}
We use the Min\_cf\_algorithm as described to find the k closest points per each cf range returned. To find the k closest points to a point $q$, we just need to search through the list of points returned by the Min\_cf\_algorithm for each counterfactual range. The proof of correctness of the Min\_cf\_algorithm is found in Theorem \ref{Min_Cf_Theorem}.

\begin{algorithm}[H]
  \caption{Min\_Cf\_algorithm}\label{Min_Cf_algorithm}
  \begin{algorithmic}[1]      
    \REQUIRE Sets $X_1,\dots,X_N$; query $\mathbf q$; integer $k$;
             distance metric $d$ ($L_0$, $L_1$, $L_2$)
    \ENSURE  $S_k$ – the $k$ points closest to $\mathbf q$

    \STATE \textbf{Step 1: Trim each dimension}
    \FOR{$i = 1$ \TO $N$}
      \STATE $X_i' \gets$ the $k$ values in $X_i$ closest to $q_i$
    \ENDFOR

    \STATE \textbf{Step 2: Form candidate points}
    \STATE $C \gets X_1' \times \dots \times X_N'$ \COMMENT{Cartesian product}

    \STATE \textbf{Step 3: Pick the best $k$}
    \FORALL{$\mathbf x \in C$}
      \STATE compute $d(\mathbf x,\mathbf q)$
    \ENDFOR
    \RETURN the $k$ points in $C$ with the smallest distance
  \end{algorithmic}
\end{algorithm}

\subsection{Intervene}
\begin{algorithm}[!ht]
\caption{\textbf{intervene}: reach a causally consistent state from a causally consistent current state}
\label{alg_intervene}
\begin{algorithmic}[1]
    \REQUIRE Causal \textit{rules} $C$, List \textit{visited\_states}, List \textit{actions\_taken}, Actions $a\in A$:
    \begin{itemize}
    
    \item Causal Action: $s$ gets altered to a causally
    consistent new state $s'=a(s)$. OR 
    \item Direct Action: new state $s'=a(s)$ is obtained by altering 1 feature value of $s$.     
    \end{itemize}

    \STATE Set $(s, actions\_taken)$ = \textit{pop(visited\_states)}
    \STATE Try to select an action $a\in A$ ensuring \textit{not\_member(a(s),visited\_states)} and \textit{not\_member(a,actions\_taken)} are $TRUE$
    \IF{ $a$ exists}
    \STATE Set $(s, actions\_taken),visited\_states=update(s,visited\_states,actions\_taken,a)$
    \ELSE
    \STATE //Backtracking 
    \IF {\textit{visited\_states} is empty }
    \STATE \textit{EXIT with Failure}
    \ENDIF
    \STATE Set $(s, actions\_taken)$ = \textit{pop(visited\_states)}
    \ENDIF
    \STATE Set $(s,actions\_taken), visited\_states=$\\ \hspace{0.2 in} $make\_consistent(s,actions\_taken, visited\_states,C,A)$ 
    \STATE Append $(s,actions\_taken)$ to \textit{visited\_states}
    \STATE Return \textit{visited\_states}.

\end{algorithmic}
\end{algorithm}

Function `\textit{\textbf{intervene}}' implements the transition function $\delta$ from Definition \ref{defn:delta}. It is called by `\textit{\textbf{find\_path}}'. The primary purpose of `\textit{\textbf{intervene}}' is to transition from the current state to the next state, ensuring actions are not repeated and states are not revisited. 
In Algorithm \ref{alg_intervene}, we specify the pseudo-code, which takes as arguments an Initial State $I$ that is causally consistent, a set of Causal Rules $C$, and a set of actions $A$. The function \textit{intervene} acts as a transition function that takes as input a list \textit{visited\_states} containing the current state $s$ as the last element, and returns the new state $s'$ by appending $s'$ to \textit{visited\_states}. The new state $s'$ is what the current state $s$ traverses. Additionally, the function \textit{intervene} ensures that no states are revisited. In traversing from $s$ to $s'$, if there are a series of intermediate states that are \textbf{not} causally consistent, it is also included in \textit{visited\_states}, thereby depicting how to traverse from 1 causally consistent state to another.

\subsection{Checking for Counterfactual/Goal State: is\_counterfactual}

\begin{algorithm}[!ht]
\caption{\textbf{is\_counterfactual}: checks if a state is a counterfactual/goal state}
\label{alg_counterfactual}
\begin{algorithmic}[1]
    \REQUIRE State $s\in S$, Set of Causal \textit{rules} $C$, Set of Decision \textit{rules} $Q$
    \IF{$s$ satisfies \textbf{ALL} rules in $C$ \textbf{AND} $s$ satisfies \textbf{NO} rules in $Q$}
    \STATE Return $TRUE$.
    \ELSE    
    \STATE Return $FALSE$.
    \ENDIF
\end{algorithmic}
\end{algorithm}

The function \textit{is\_counterfactual} is our algorithmic implementation of checking if a state $s\in G$ from definition \ref{defn:G}.
In Algorithm \ref{alg_counterfactual}, we specify the pseudo-code for a function \textit{is\_counterfactual} which takes as arguments a state $s\in S$, a set of causal rules $C$, and a set of Decision rules $Q$. The function checks if a state $s\in S$ is a counterfactual/goal state. By definition \textit{is\_counterfactual} is $TRUE$ for state $s$ that is causally consistent with all $c\in C$ and \textbf{does not} agree with the any decision rules $q\in Q$.

\vspace{-0.15 in}
\begin{equation}
   is\_counterfactual(s,C,Q)=TRUE \mid s\ agrees\ with\ C;\ s\ disagrees\ with\ Q; 
\end{equation}

\subsection{Make Consistent}
\begin{algorithm}[!ht]
\caption{\textbf{make\_consistent}: reaches a consistent state}
\label{alg_inner_delta}
\begin{algorithmic}[1]
    \REQUIRE State $s$, Causal \textit{rules} $C$, List \textit{visited\_states} , \textit{actions\_taken}, Actions $a\in A$:
    \WHILE{$s$ does not satisfy all rules in $C$}
    \STATE Try to select a causal action $a$ ensuring \textit{not\_member(a(s),visited\_states)} and \textit{not\_member(a,actions\_taken)} are $TRUE$
    \IF{ causal action $a$ exists}
    \STATE Set \textit{(s,actions\_taken),visited\_states=update(s,visited\_states,actions\_taken,a)}
    \ELSE
        \STATE Try to select a direct action $a$ ensuring \textit{not\_member(a(s),visited\_states)} and \textit{not\_member(a,actions\_taken)} are $TRUE$
        \IF{ direct action $a$ exists}
        \STATE Set \textit{(s, actions\_taken), visited\_states=update(s,visited\_states,actions\_taken,a)}
        \ELSE
        \STATE //Backtracking 
        \IF {\textit{visited\_states} is empty } 
        \STATE \textit{EXIT with Failure}
        \ENDIF
        \STATE Set $(s, actions\_taken)$ = \textit{pop(visited\_states)}
        \ENDIF
    \ENDIF
    \ENDWHILE
    \STATE Return $(s,actions\_taken),visited\_states$ .

\end{algorithmic}
\end{algorithm}

The pseudo-code for `\textit{\textbf{make\_consistent}}' is specified in Algorithm \ref{alg_inner_delta}. It takes as arguments a current State $s$, a list \textit{actions\_taken}, a list \textit{visited\_states}, a set of Causal Rules $C$ and a set of actions $A$. Called by `\textit{\textbf{intervene}}', `\textit{\textbf{make\_consistent}}' transitions from the current state to a new, causally consistent state.

\subsection{Update}\label{sec_update}
\begin{algorithm}[!ht]
\caption{\textbf{update}: Updates the list \textit{actions\_taken} with the planned action. Then updates the current state.}
\label{alg_update}
\begin{algorithmic}[1]
    \REQUIRE State $s$, List \textit{visited\_states}, List \textit{actions\_taken}, Action $a \in A$:
    \begin{itemize}
        \item Causal Action: $s$ gets altered to a causally
        consistent new state $s'=a(s)$. OR 
        \item Direct Action: new state $s'=a(s)$ is obtained by altering 1 feature value of $s$.  
    \end{itemize}
    \STATE Append $a$ to \textit{actions\_taken}.
    \STATE Append $(s, actions\_taken)$ to \textit{visited\_states}.
    \STATE Set $s=a(s)$.
    \RETURN $(s, [~])$, \textit{visited\_states}

\end{algorithmic}
\end{algorithm}

Function `\textit{\textbf{update}}' tracks the list of actions taken and states visited to avoid repeating actions and revisiting states. In Algorithm \ref{alg_update}, we specify the pseudo-code for the \textit{update} function, that given a state $s$, list \textit{actions\_taken}, list \textit{visited\_states}and given an action $a$, appends $a$ to \textit{actions\_taken}. It also appends the list  \textit{actions\_taken} as well as the new resultant state resulting from the action $a(s)$ to the list  \textit{visited\_states}. The list \textit{actions\_taken} is used to track all the actions attempted from the current state to avoid repeating them. The function \textit{update} is called by both functions \textit{intervene} and \textit{make\_consistent}.

\section{Proofs}
\subsection*{Min\_Cf\_ Theorem}
We claim that the 
Proof of Correctness of the Algorithm \ref{Min_Cf_algorithm}.

\subsubsection*{Statement}
Let \( X_1, X_2, \ldots, X_N \) be sets of real numbers, and consider the Cartesian product:

\[
\mathcal{X} = X_1 \times X_2 \times \cdots \times X_N.
\]

Given a query point \( \mathbf{q} = (q_1, q_2, \ldots, q_N) \in \mathbb{R}^N \) that is not necessarily in \( \mathcal{X} \), and a distance metric \( d \) (e.g., \( L_1 \), \( L_2 \), or \( L_0 \)), let \( S_k \subseteq \mathcal{X} \) denote the set of the \( k \) points in \( \mathcal{X} \) closest to \( \mathbf{q} \) under \( d \).

Let \( X_i' \subseteq X_i \) be the set of \( k \) values in \( X_i \) closest to \( q_i \), for each \( i = 1, \ldots, N \). Define:

\[
\mathcal{X}' = X_1' \times X_2' \times \cdots \times X_N'.
\]

Then:

\[
S_k \subseteq \mathcal{X}'.
\]

\subsubsection*{Proof (for \( L_0 \) , \( L_1 \) and \( L_2 \))} \label{Min_Cf_Theorem}

Assume, for the sake of contradiction, that there exists a point \( \mathbf{x} = (x_1, x_2, \ldots, x_N) \in S_k \) such that \( \mathbf{x} \notin \mathcal{X}' \). Then, there exists a non-empty index set \( J \subseteq \{1, \ldots, N\} \) such that for all \( j \in J \), we have \( x_j \notin X_j' \).

By definition of \( X_j' \), this means that for each \( j \in J \), there exist at least \( k \) elements \( x_j^{(1)}, x_j^{(2)}, \ldots, x_j^{(k)} \in X_j' \) such that:

\[
|x_j^{(m)} - q_j| < |x_j - q_j| \quad \text{for all } m = 1, \ldots, k.
\]

For each \( j \in J \), choose an index \( m_j \in \{1, \ldots, k\} \) and define \( x_j^{(m_j)} \in X_j' \) as one of the values strictly closer to \( q_j \) than \( x_j \). This gives us a selected replacement value for each coordinate \( j \in J \).

We now construct a new point \( \mathbf{x}' = (x_1', x_2', \ldots, x_N') \in \mathcal{X} \) where:

\[
x_i' = \begin{cases}
x_i, & i \notin J \\
x_i^{(m_i)}, & i \in J
\end{cases}
\quad \text{with } x_i^{(m_i)} \in X_i' \text{ and } |x_i^{(m_i)} - q_i| < |x_i - q_i|.
\]

Now consider the distance from \( \mathbf{x}' \) to \( \mathbf{q} \).

\subsection*{Case 1: \( L_1 \) distance}

\[
d_1(\mathbf{x}', \mathbf{q}) = \sum_{i=1}^N |x_i' - q_i| = \sum_{i \notin J} |x_i - q_i| + \sum_{i \in J} |x_i^{(m_i)} - q_i| < \sum_{i \notin J} |x_i - q_i| + \sum_{i \in J} |x_i - q_i| = d_1(\mathbf{x}, \mathbf{q}).
\]

\subsection*{Case 2: \( L_2 \) distance}

\[
d_2(\mathbf{x}', \mathbf{q}) = \sqrt{ \sum_{i=1}^N (x_i' - q_i)^2 } = \sqrt{ \sum_{i \notin J} (x_i - q_i)^2 + \sum_{i \in J} (x_i^{(m_i)} - q_i)^2 } < \]

\[\sqrt{ \sum_{i \notin J} (x_i - q_i)^2 + \sum_{i \in J} (x_i - q_i)^2 } = d_2(\mathbf{x}, \mathbf{q}).
\]

\subsection*{Case 3: \( L_0 \) distance}

The \( L_0 \) distance counts the number of coordinates in which two vectors differ:

\[
d_0(\mathbf{x}, \mathbf{q}) = \sum_{i=1}^N \mathbf{1}_{x_i \ne q_i}.
\]

Assume, for the sake of contradiction, that there exists a point \( \mathbf{x} = (x_1, \ldots, x_N) \in S_k \) such that \( \mathbf{x} \notin \mathcal{X}' \). Then there exists a non-empty index set \( J \subseteq \{1, \ldots, N\} \) such that for all \( j \in J \), we have \( x_j \notin X_j' \).

For each \( j \in J \), let \( x_j^{(m_j)} \in X_j' \) be one of the values closer to \( q_j \) than \( x_j \). Define the point \( \mathbf{x}' = (x_1', \ldots, x_N') \in \mathcal{X} \) as:

\[
x_i' = \begin{cases}
x_i^{(m_i)}, & i \in J \\
x_i, & i \notin J
\end{cases}
\quad \text{with } x_i^{(m_i)} \in X_i'.
\]

Then, since \( x_i^{(m_i)} \) is closer to \( q_i \) than \( x_i \), two cases arise:

\begin{itemize}
  \item If \( x_i^{(m_i)} = q_i \), then the contribution to \( d_0 \) from coordinate \( i \) is 0 (improvement).
  \item Now, if \( x_i^{(m_i)} \ne q_i \), then the contribution to \( d_0 \) from coordinate \( i \) remains 1. (Note: Since $x_i$ is further away from $q_i$ than \( x_i^{(m_i)} \), its original contribution to \( d_0 \) was already 1 since $x_i \ne q_i$.)
\end{itemize}

Thus, the overall L0 distance does not increase.

\[
d_0(\mathbf{x}', \mathbf{q}) \le d_0(\mathbf{x}, \mathbf{q})
\]

This contradicts the assumption that \( \mathbf{x} \in S_k \), so we conclude:

\[
S_k \subseteq \mathcal{X}'.
\]

\hfill\(\blacksquare\)

\begin{theorem}{Soundness Theorem}\label{theorem_soundness}\\
\noindent Given a CFG $\mathbb{X}=(S_C,S_Q,I,\delta)$, constructed from a run of algorithm \ref{alg_path} and a corresponding candidate path $P$, $P$ is a solution path for $\mathbb{X}$.
\begin{proof}
Let $G$ be a goal set for $\mathbb{X}$. By definition \ref{defn:candidate_path} $P=s_0,...,s_{m}$, where $m\geq0$.
By definition \ref{solution_path}
we must show $P$ has the following properties.
        
        1) $s_0=I $
        
        2) $s_m\in G $
        
        3) $s_j\in S_C\ for\ all\ j\ \in\{0,...,m\}$
        
        4) $s_0,...,s_{m-1} \not\in G$
        
        5) $s_{i+1}\in \delta(s_i)\ for\ i\ \in\{0,...,m-1\}$\\
1) By definition \ref{defn:I}, $I$ is causally consistent and cannot be removed from the candidate path. Hence I must be in the candidate path and is the first state as per line 2 in algorithm \ref{alg_path}. Therefore $s_0$ must be $I$.\\
2) The while loop in algorithm 5 ends if and only if $is\_counterfactual(s,C,Q)$ is True. From theorem 1 $is\_counterfactual(s,C,Q)$ is True only for the goal state. Hence $s_m\in G$.\\
3) By  definition\ref{defn:candidate_path} of the candidate path, all states $s_j\in S_C\ for\ all\ j\ \in\{0,...,m\}$.\\
4) By theorem \ref{theorem_all_but_last}, we have proved the claim $s_0,...,s_{m-1} \not\in G$.\\
5) By theorem \ref{theorem_delta}, we have proved the claim $s_{i+1}\in \delta(s_i)\ for\ i\ \in\{0,...,m-1\}$.\\
Hence we proved the candidate path $P$ (definition \ref{defn:candidate_path}) is a solution path (definition \ref{solution_path}).







\end{proof}
\end{theorem}



\begin{theorem}\label{theorem_is_counterfactual}
\noindent Given a CFG $\mathbb{X}=(S_C,S_Q,I,\delta)$, constructed from a run of algorithm \ref{alg_path}, with goal set $G$, and $s\in S_C$; $is\_counterfactual(s,C,Q)$ will be $TRUE$ if and only if $s\in G$.
\begin{proof}

By the definition of the goal set $G$ we have
\begin{equation}
G = \{ s \in S_C| s\not\in S_Q\} \label{theorem_1_1}
\end{equation}
For $is\_counterfactual$ which takes as input the state $s$, the set of causal rules $C$ and the set of decision rules $Q$ (Algorithm \ref{alg_counterfactual}), we see that by from line 1 in algorithm \ref{alg_counterfactual}, it returns TRUE if it satisfied all rules in $C$ and no rules in $Q$.

By the definition \ref{defn:S_Q}, $s\in S_Q$ \textit{if and only if} it satisfies a rule in $Q$. 
Therefore, $is\_counterfactual(s,C,Q)$ is $TRUE$ if and only if $s\not\in S_Q$ and since $s\in S_C$ and $s\not\in S_Q$ then $s\in G$.

\end{proof}
\end{theorem}

\begin{theorem}\label{theorem_delta}
\noindent Given a CFG $\mathbb{X}=(S_C,S_Q,I,\delta)$, constructed from a run of algorithm \ref{alg_path} and a corresponding candidate path $P=s_0,...,s_{m}$; $s_{i+1}\in \delta(s_i)\ for\ i\ \in\{0,...,m-1\}$

\begin{proof}
This property can be proven by induction on the length of the list \textit{visited\_lists} obtained from Algorithm 5,4,3.\\
\textbf{Base Case}: The list \textit{visited\_lists} from algorithm \ref{alg_path} has length of 1, i.e., [$s_0$]. The property $s_{i+1}\in \delta(s_i)\ for\ i\ \in\{0,...,m-1\}$ is trivially true as there is no $s_{-1}$.\\
\textbf{Inductive Hypotheses}: 
We have a list [$s_0,...,s_{n-1}$] of length $n$ generated from $0$ or more iteration of running the function \textit{intervene} (algorithm \ref{alg_intervene}), and it  satisfies the claim $s_{i+1}\in \delta(s_i)\ for\ i\ \in\{0,...,n-1\}$\\

\textbf{Inductive Step}: If we have a list  [$s_0,...,s_{n-1}$]  of length n and we wish to get element $s_n$ obtained through running another iteration of function \textit{intervene} (algorithm \ref{alg_intervene}). Since [$s_0,...,s_{n-1}$] is  of length n by the inductive hypothesis, it satisfies the property, and it is sufficient to show $s_n\in\delta(s_{n-1})$ where $s_{i+1}\in \delta(s_i)\ for\ i\ \in\{0,...,n-1\}$.\\

The list \textit{visited\_lists} from algorithm \ref{alg_path} has length of $n$. Going from $s_{n-1}$ to $s_n$ involves calling the function \textit{intervene} (algorithm \ref{alg_intervene}) which in turn calls the function \textit{make\_consistent} (algorithm \ref{alg_inner_delta}).

Function \textit{make\_consistent} (algorithm \ref{alg_inner_delta}) takes as input the state $s$, the list of actions taken \textit{actions\_taken}, the list of visited states \textit{visited\_states}, the set of causal rules $C$ and the set of possible actions $A$.
It returns \textit{visited\_states} with the new causally consistent states as the last element. From line 1, if we pass as input a causally consistent state, then function \textit{make\_consistent} does nothing. On the other hand, if we pass a causally inconsistent state, it takes actions to reach a new state. Upon checking if the action taken results in a new state that is causally consistent from the \textit{while} loop in line 1, it returns the new state. 
Hence, we have shown that the moment a causally consistent state is encountered in function \textit{make\_consistent}, it does not add any new state.

Function \textit{intervene} (algorithm \ref{alg_intervene}) takes as input the list of visited states \textit{visited\_states} which contains the current state as the last element, the set of causal rules $C$ and the set of possible actions $A$. It returns \textit{visited\_states} with the new causally consistent states as the last element. It calls the function \textit{make\_consistent}. For the function \textit{intervene}, in line 1 it obtains the current state (in this case $s_{n-1}$) from the list \textit{visited\_states}. It is seen in line 2 that an action $a$ is taken: 

        1) Case 1: If a causal action is taken, then upon entering the the function \textit{make\_consistent} (algorithm \ref{alg_inner_delta}), it will not do anything as causal actions by definition result in causally consistent states.         

        2) Case 2: If a direct action is taken, then the new state that may or may not be causally consistent is appended to \textit{visited\_states}. The call to the function \textit{make\_consistent} will append one or more states with only the final state appended being causally consistent.

Hence we have shown that the moment a causally consistent state is appended in function \textit{intervene}, it does not add any new state. This causally consistent state is $s_n$. In both cases $s_n=\sigma(s_{n-1})$ as defined in definition \ref{defn:impl_cfg} and this $s_n \in \delta(s_{n-1})$.




\end{proof}
\end{theorem}

\begin{theorem}\label{theorem_all_but_last}
\noindent Given a CFG $\mathbb{X}=(S_C,S_Q,I,\delta)$, constructed from a run of algorithm \ref{alg_path}, with goal set $G$ and a corresponding candidate path $P=s_0,...,s_{m}$ with $m\geq0$, $s_0,...,s_{m-1} \not\in G$.
\begin{proof}
This property can be proven by induction on the length of the list \textit{visited\_lists} obtained from Algorithm 5,4,3.\\
\textbf{Base Case}: \textit{visited\_lists} has length of 1. 
Therefore the property $P=s_0,...,s_{m}$ with $m\geq0$, $s_0,...,s_{m-1} \not\in G$ is trivially true as state $s_{j}$ for $j<0$ does not exist.\\


\textbf{Inductive Hypotheses}: 
We have a list [$s_0,...,s_{n-1}$] of length $n$ generated from $0$ or more iteration of running the function \textit{intervene} (algorithm \ref{alg_intervene}), and it  satisfies the claim $s_0,...,s_{n-2} \not\in G$.\\

\textbf{Inductive Step}: Suppose we have a list [$s_0,...,s_{n-1}$] of length n and we wish to append the n+1 th element (state $s_{n}$) by calling the function \textit{intervene}, and we wish to show that that the resultant list satisfies the claim $s_0,...,s_{n-1} \not\in G$. The first n-1 elements ($s_0,...,s_{n-2}$) are not in $G$ as per the inductive hypothesis.

From line 3 in the function \textit{get\_path} (algorithm \ref{alg_path}), we see that to call the function \textit{intervene} another time, the current state (in this case $s_{n-1})$ \textbf{cannot} be a counterfactual, by theorem \ref{theorem_is_counterfactual}. Hence $s_{n-1}\not\in G$ 

Therefore by induction the claim $s_0,...,s_{n-1}\not \in G$ holds.
\end{proof}
\end{theorem}
\newpage
\section{Experiments}
\subsection{Tables from Experiments}
\begin{table}[htbp]
\centering
\caption{Performance on \textbf{Adult}, \textbf{German Credit} and \textbf{Car Evaluation} Datasets}
\label{tbl:adult_german}
\resizebox{\linewidth}{!}{%
\begin{tabular}{@{}l l c c c c c@{}}
\toprule
\thead{Dataset} & \thead{Model} & \thead{Fid.\\(\%)} & 
\thead{Acc.\\(\%)} & \thead{Prec.\\(\%)} & 
\thead{Rec.\\(\%)} & \thead{F1\\(\%)} \\ \midrule
\multirow{8}{*}{Adult} 
  & DNN                 & N/A & 85.57$\pm$0.38 & 85.0$\pm$0.63 & 85.8$\pm$0.40 & 85.0$\pm$0.63 \\
  & FOLD-SE[DNN]        & 93.16$\pm$0.60 & 84.2$\pm$0.28 & 83.4$\pm$0.49 & 84.2$\pm$0.40 & 83.4$\pm$0.49 \\ \cmidrule{2-7}
  & GBC                 & N/A & 86.45$\pm$0.33 & 85.8$\pm$0.40 & 86.6$\pm$0.49 & 85.8$\pm$0.40 \\
  & FOLD-SE[GBC]        & 95.94$\pm$0.67 & 85.24$\pm$0.23 & 84.6$\pm$0.49 & 85.2$\pm$0.40 & 84.2$\pm$0.40 \\ \cmidrule{2-7}
  & RF                  & N/A & 85.60$\pm$0.27 & 85.0$\pm$0.00 & 85.6$\pm$0.49 & 85.2$\pm$0.40 \\
  & FOLD-SE[RF]         & 90.27$\pm$0.30 & 84.37$\pm$0.23 & 83.4$\pm$0.49 & 84.4$\pm$0.49 & 83.2$\pm$0.40 \\ \cmidrule{2-7}
  & LR                  & N/A & 84.78$\pm$0.28 & 84.2$\pm$0.40 & 84.8$\pm$0.40 & 84.2$\pm$0.40 \\
  & FOLD-SE[LR]         & 94.03$\pm$0.36 & 83.92$\pm$0.32 & 83.0$\pm$0.00 & 83.8$\pm$0.40 & 83.4$\pm$0.49 \\ \midrule
\multirow{8}{*}{German} 
  & DNN                 & N/A & 74.5$\pm$1.67 & 73.0$\pm$2.19 & 74.4$\pm$1.74 & 73.0$\pm$2.19 \\
  & FOLD-SE[DNN]        & 81.5$\pm$2.32 & 71.6$\pm$1.50 & 71.0$\pm$2.37 & 71.6$\pm$1.50 & 70.8$\pm$1.72 \\ \cmidrule{2-7}
  & GBC                 & N/A & 75.8$\pm$1.63 & 74.6$\pm$1.62 & 76.0$\pm$1.67 & 74.6$\pm$1.62 \\
  & FOLD-SE[GBC]        & 81.6$\pm$5.42 & 72.6$\pm$3.64 & 72.0$\pm$2.90 & 72.6$\pm$3.61 & 71.0$\pm$3.03 \\ \cmidrule{2-7}
  & RF                  & N/A & 75.7$\pm$1.21 & 74.6$\pm$2.15 & 76.0$\pm$1.41 & 72.8$\pm$1.17 \\
  & FOLD-SE[RF]         & 85.1$\pm$2.37 & 71.6$\pm$0.20 & 70.2$\pm$1.94 & 71.2$\pm$0.40 & 66.2$\pm$2.86 \\ \cmidrule{2-7}
  & LR                  & N/A & 74.5$\pm$1.18 & 73.0$\pm$1.26 & 74.6$\pm$1.36 & 73.4$\pm$1.02 \\
  & FOLD-SE[LR]         & 82.5$\pm$2.41 & 72.2$\pm$2.01 & 71.4$\pm$2.87 & 72.0$\pm$2.19 & 71.8$\pm$2.48 \\ \midrule
  
  \multirow{8}{*}{Car}
  & DNN          & N/A               & 94.4$\pm$0.31 & 97.6$\pm$0.49 & 97.2$\pm$0.40 & 97.2$\pm$0.40 \\
  & FOLD-SE[DNN] & 91.55$\pm$3.95    & 91.6$\pm$4.00 & 93.6$\pm$2.50 & 91.6$\pm$3.88 & 91.8$\pm$3.87 \\ \cmidrule{2-7}
  & GBC          & N/A               & 97.5$\pm$1.12 & 97.4$\pm$1.02 & 97.4$\pm$1.02 & 97.4$\pm$1.02 \\
  & FOLD-SE[GBC] & 97.16$\pm$3.80    & 95.24$\pm$4.30 & 96.4$\pm$2.80 & 95.4$\pm$4.27 & 95.4$\pm$4.27 \\ \cmidrule{2-7}
  & RF           & N/A               & 95.71$\pm$0.56 & 95.6$\pm$0.80 & 95.6$\pm$0.80 & 95.6$\pm$0.80 \\
  & FOLD-SE[RF]  & 94.27$\pm$3.27    & 95.08$\pm$2.92 & 96.0$\pm$2.19 & 95.2$\pm$3.12 & 95.4$\pm$2.87 \\ \cmidrule{2-7}
  & LR           & N/A               & 94.79$\pm$1.43 & 95.0$\pm$1.79 & 94.8$\pm$1.72 & 94.8$\pm$1.72 \\
  & FOLD-SE[LR]  & 95.37$\pm$1.21    & 94.32$\pm$1.54 & 94.8$\pm$1.17 & 94.4$\pm$1.50 & 94.4$\pm$1.50 \\ \bottomrule
\end{tabular}}
\end{table}

\section{Experimental Setup}

\subsubsection{Dataset: Adult }
We run the FOLD-SE algorithm to produce the following decision making rules: 

    {\tt label(X,'<=50K') :- not marital\_status(X,'Married-civ-spouse')\\
    \indent\indent\indent \indent\indent\indent , capital\_gain(X,N1), N1=<6849.0.}

    {\tt label(X,'<=50K') :- marital\_status(X,'Married-civ-spouse')\\
    \indent\indent\indent \indent\indent\indent , capital\_gain(X,N1), N1=<5013.0, education\_num(X,N2), N2=<12.0.}

 \begin{enumerate}
     \item Accuracy: 84.5\%
     \item Precision: 86.5\%
     \item Recall: 94.6\%
 \end{enumerate}

2) FOLD-SE gives Causal rules for the `marital\_status' feature having  value `never\_married':

    {\tt marital\_status(X,'Never-married') :- not relationship(X,'Husband'), 
    
    \indent\indent\indent\indent\indent\indent\indent\indent\indent\indent not relationship(X,'Wife'), age(X,N1), N1=<29.0.} 

 \begin{enumerate}
     \item Accuracy: 86.4\%
     \item Precision: 89.2\%
     \item Recall: 76.4\%
 \end{enumerate}

3) FOLD-SE gives Causal rules for the `marital\_status' feature having  value `Married-civ-spouse':

    {\tt marital\_status(X,'Married-civ-spouse') :- relationship(X,'Husband').} 

    {\tt marital\_status(X,'Married-civ-spouse') :- relationship(X,'Wife').} 

 \begin{enumerate}
     \item Accuracy: 99.1\%
     \item Precision: 99.9\%
     \item Recall: 98.2\%
 \end{enumerate}

4) For values of the feature `marital\_status' that are not `Married-civ-spouse'  or `never\_married' which we shall call `neither', a user defined rule is used
    
    {\tt marital\_status(X,neither) :- not relationship(X,'Husband')\\
    \indent\indent\indent \indent\indent\indent , not relationship(X,'Wife').}

5) FOLD-SE gives Causal rules for the `relationship' feature having  value `husband':

    {\tt relationship(X,'Husband') :- not sex(X,'Male'), age(X,N1), not(N1=<27.0).} 

 \begin{enumerate}
     \item Accuracy: 82.3\%
     \item Precision: 71.3\%
     \item Recall: 93.2\%
 \end{enumerate}
 
5) For the  `relationship' feature value of `wife', a user defined rule is used
    
    {\tt relationship(X,'Wife') :- sex(X,'Female').} 
    
6)Features Used in Generating the counterfactual path:
\begin{itemize}
    \item Feature: marital\_status
    \item Feature: relationship
    \item Feature: sex

    \item capital\_gain
    \item education\_num
    \item age
    
\end{itemize}

\subsubsection{Dataset: German }
We run the FOLD-SE algorithm to produce the following decision making rules: 

    {\tt label(X,'good') :- checking\_account\_status(X,'no\_checking\_account')}

    {\tt label(X,'good') :- not checking\_account\_status(X,'no\_checking\_account')  \indent\indent\indent\indent\indent\indent , not credit\_history(X,'all\_dues\_atbank\_cleared') \indent\indent\indent\indent\indent\indent , duration\_months(X,N1), N1=<21.0, credit\_amount(X,N2)
    \indent\indent\indent\indent\indent\indent , not(N2=<428.0), not ab1(X,'True').}

    {\tt ab1(X,'True') :- property(X,'car or other')\\
    \indent\indent\indent \indent\indent\indent , credit\_amount(X,N2), N2=<1345.0.}
    
 \begin{enumerate}
     \item Accuracy: 77\%
     \item Precision: 83\%
     \item Recall: 84.2\%
 \end{enumerate}

2) FOLD-SE gives Causal rules for the `present\_employment\_since' feature having  value `employed' where employed is the placeholder for all feature values that are \textbf{not} equal to the feature value `unemployed':

    {\tt present\_employment\_since(X,'employed') :- \\
    \indent\indent\indent\indent\indent\indent not job(X,'unemployed/unskilled-non\_resident').

 \begin{enumerate}
     \item Accuracy: 95\%
     \item Precision: 96.4\%
     \item Recall: 98.4\%
 \end{enumerate}

3) For values of the feature `present\_employment\_since' that are `unemployed', a user defined rule is used

    {\tt present\_employment\_since(X,'unemployed') :- \\
    \indent\indent\indent\indent\indent\indent job(X,'unemployed/unskilled-non\_resident').

6)Features Used in Generating the counterfactual path:
\begin{itemize}
    \item checking\_account\_status
    
    \item credit\_history
    
    \item property

    \item duration\_months
    
    \item credit\_amount
    
    \item present\_employment\_since

    \item job 
\end{itemize}

\subsubsection{Dataset: Cars }

We run the FOLD-SE algorithm to produce the following rules: 

    {\tt label(X,'negative') :- persons(X,'2').}

    {\tt label(X,'negative') :- safety(X,'low').}

    {\tt label(X,'negative') :- buying(X,'vhigh'), maint(X,'vhigh').}

    {\tt label(X,'negative') :- not buying(X,'low'), not buying(X,'med'), maint(X,'vhigh').}

    {\tt label(X,'negative') :- buying(X,'vhigh'), maint(X,'high').}

The rules described above indicate if the purchase of a car was rejected . 
 \begin{enumerate}
     \item Accuracy: 93.9\%
     \item Precision: 100\%
     \item Recall: 91.3\%
 \end{enumerate}

2) Features and Feature Values used:
\begin{itemize}
    \item Feature: persons
    \item Feature: safety
    \item Feature: buying
    \item Feature: maint

\end{itemize}

\end{document}